\documentclass[lettersize,journal]{IEEEtran}

\usepackage{amsmath}
\usepackage{amssymb}
\usepackage{mathtools}
\usepackage{amsthm}
\usepackage[T1]{fontenc}
\usepackage[utf8]{inputenc}
\usepackage{microtype}
\usepackage{graphicx}
\usepackage{booktabs}
\usepackage{hyperref,comment}
\usepackage{tikz}
\usetikzlibrary{arrows.meta,positioning,shapes.geometric,calc}
\newtheorem{definition}{Definition}
\newtheorem{theorem}{Theorem}
\newtheorem{proposition}{Proposition}
\newtheorem{corollary}{Corollary}
\newtheorem{remark}{Remark}

\newcommand{\im}{\operatorname{im}}
\newcommand{\twr}{\operatorname{twr}}

\usepackage[capitalize,noabbrev]{cleveref}

\title{Persistent Homology as a Theory of Emergent Structure}

\author{%
  Xin Li,\textsuperscript{1}\thanks{This work was partially supported by NSF IIS-2401748 and BCS-2401398. The author has used Claude 4.8 and ChatGPT 5.5 models to assist in the development of theoretical ideas, mathematical proofs, and visual illustrations presented in this paper.} \\
  Department of Computer Science\\
  University at Albany\\
  \texttt{xli48@albany.edu}
}
\begin{document}

\maketitle

\begin{abstract}
Why do some macroscopic structures remain identifiable even though their
microscopic constituents continually change? Vortices persist while fluid
parcels turn over, neural memories persist while spikes and synapses fluctuate,
and institutions persist while individuals enter and leave. We propose a
scale-relative answer: an emergent property is a persistent nontrivial homology
class $[z]\in H_p=\ker\partial_p/\im\partial_{p+1}$, a macro-feature that is
closed but not exact across a filtration of descriptions. This identification
turns emergence into a \emph{measurement} problem. Persistent bars detect stable
macro-features, and we introduce a contractive-similarity (CS) graph operator to supply scaffold spectral gaps that predict robustness. Hodge decomposition separates harmonic macro-scaffold from exact and co-exact micro-flow; and functorial
condensation explains when one level's emergent class becomes a unit for the
next. The resulting scaffold-flow framework expresses six
familiar signatures of emergence (i.e., inevitability, coherence, irreducibility,
complementarity, robustness, and hierarchy) within one mathematical language.
It also yields falsifiable predictions across atmospheric, neural, and social
systems: genuine emergent structures should persist across filtrations, remain
spectrally stable, respond disproportionately to harmonic interventions, and
require timescale separation for hierarchical autonomy.
\end{abstract}

\section{Introduction}
\label{sec:0}

Why do some macroscopic structures remain identifiable while their microscopic
constituents continually change? Vortex filaments persist through fluid
turnover~\cite{she1990intermittent}; cell types maintain identity while
molecular components are replaced~\cite{waddington1957strategy}; institutions
survive the entry and exit of individuals~\cite{north1990institutions}; and
neural attractors remain stable despite fluctuating spikes and synapses
\cite{hopfield1982neural}. These examples suggest that emergence is not simply
aggregation, but the principled formation of a relational scaffold that can persist
through micro-level turnover.

We argue that the persistence-through-turnover problem is central to emergence. Across
statistical physics, biology, neuroscience, and the social sciences,
emergence names the appearance of coherent macroscopic order from many
interacting microscopic components. However, the concept remains more evocative
than precise. The philosophical literature~\cite{broad1925mind,kim1999emergence}
distinguishes ``weak'' emergence~\cite{bedau1997weak}, in which the macro-level
is a useful summary of micro-dynamics, from ``strong'' emergence
\cite{chalmers2006emergence}, in which the macro-level is genuinely novel. What
is still missing is a mathematical object that makes the distinction
operational.

Two classical insights clarify what such an object must capture. In Anderson's
formulation \cite{anderson1972more}, ``more is different'': new levels of organization require new
concepts, with symmetry breaking as the paradigmatic route by which symmetric
microscopic rules produce ordered macroscopic states
\cite{weinberg1976implications}. Simon adds the complementary
organizational principle that complex systems are often hierarchical and nearly
decomposable \cite{simon1962architecture}: stable macro-structures arise when internal couplings within
subsystems are stronger than couplings between subsystems. Together, these views suggest that emergence is
neither arbitrary novelty nor naive aggregation, but stable global organization
that is coherent across boundaries, irreducible to local description, and
available as a unit for higher-level organization \cite{waldrop1993complexity}.

In this paper, we identify such an object with a \emph{nontrivial homology
class} of a persistent structure \cite{hatcher2002algebraic}: a class
$[z]\in H_p=\ker\partial_p/\im\partial_{p+1}$
that is closed, $\partial z=0$, and self-consistent, but not exact,
$[z]\neq0$, and not the boundary of lower-level parts. In this sense,
``more than the sum of its parts'' means that the emergent component lies in
the harmonic subspace of the combinatorial Hodge decomposition \cite{lim2015hodge}, orthogonal to
within-part variation. Importantly, our technical contribution is a novel \emph{measurement principle}: emergent structure can be detected by
persistent bars, scaffold spectral gaps, harmonic components, and timescale
separation. The six familiar signatures of emergence can be expressed in
one metric-topological (scaffold-flow) language, from combinatorial availability to recursive
hierarchy.

The same language also suggests how emergent structure forms and changes.
Repeated micro-level co-activation can write edges and higher-order simplices
into a scaffold, generalizing the Hebbian idea that correlations strengthen
relations \cite{hebb1949organization,curto2008cell,giusti2015clique}. Along a
filtration, qualitative change is detected by births or deaths of persistent
classes, so the Betti vector $(\beta_0,\beta_1,\beta_2,\ldots)$ becomes a
topological order parameter
\cite{edelsbrunner2002topological,kahle2014topology,linial2016phase,bobrowski2018topology}.
Unlike classical order parameters in statistical physics \cite{nakahara2003geometry,landau2013statistical}, which are often
specified in advance as scalar observables tied to symmetry breaking, our
topological order parameter is \emph{multicomponent} and \emph{scale-relative}: it records
which global relational structures appear, disappear, and persist across
levels of description.

\paragraph{What the identification is meant to do.}
The proposed definition is deliberately \emph{relative} rather than absolute.
A property is emergent not because it is mysterious, mental, or causally
unusual, but because it is nontrivial with respect to a chosen substrate,
boundary operator, and scale filtration. Changing the substrate or the
filtration can change the homology group \cite{hatcher2002algebraic}, as well as what
counts as emergent. We argue that \emph{the relativity is a feature}: it turns the vague question ``is this property emergent?'' into the operational question
``with respect to which complex, boundary map, and scale range does this
property persist as a nontrivial class?'' This scale relativity does
not replace domain modeling but supplies a common mathematical
language \cite{nakahara2003geometry} in which domain-specific models of vortices, cell types,
neural assemblies, or social institutions can be compared.

The central distinction is therefore \emph{graded}, as shown in Fig. \ref{fig:1}. An aggregate is any chain, a
coherent macro-pattern is a cycle $z\in\ker\partial_p$, and an emergent
macro-pattern is a nonzero homology class $[z]\neq0$. Persistence adds the
stability requirement that the class survive across a nontrivial filtration
interval rather than appear at a single arbitrary resolution
\cite{cohensteiner2007stability,chazal2016structure,fasy2014confidence}.
We first introduce the necessary chain-complex and CS-Laplacian machinery
and develop the six signatures as formal results, with proofs collected in the
Appendix. We then use atmospheric, neural, and social systems as application
domains, before turning to phase transitions, formation mechanisms, operational
predictions, scope, and implications.

\section{Theoretical Framework}


A \emph{chain complex} $(C_\bullet, \partial)$ is a sequence of vector
spaces $C_p$ with linear boundary maps $\partial_p: C_p \to C_{p-1}$
satisfying $\partial_p\partial_{p+1}=0$. The $p$-th homology
$H_p := \ker\partial_p / \im\partial_{p+1}$ counts the $p$-dimensional
features (components, loops, voids) that are closed but not filled in~\cite{hatcher2002algebraic};
$\beta_p := \dim H_p$ is the $p$-th Betti number. A nontrivial class
$[z]\neq 0$ in $H_p$ is a feature that no aggregation of $(p{+}1)$-dimensional
parts can produce: it is, on the present reading, an emergent feature of
degree~$p$.

\subsection*{The contractive-similarity (CS) operator: a scaffold detector}
A definition of emergence requires an operational scaffold on which cycles,
boundaries, and persistent classes can be measured. We introduce the
\emph{contractive-similarity} (CS) operator as a graph-Laplacian construction
for this purpose. The term is used here for the proposed scaffold detector, not
for a previously standardized operator. Its ingredients are familiar:
bilateral filtering couples spatial proximity with range similarity, while
spectral clustering and graph signal processing use graph Laplacians to expose
coherent low-frequency structure~\cite{tomasi1998bilateral,vonluxburg2007tutorial,ortega2018graph}.
The CS construction combines these ideas by retaining edges only when elements
are both nearby, or strongly interacting, and compatible in state, label, output,
or predicted response. It contracts away locally adjacent, but
state-incompatible variation and leaves a graph whose low-frequency structure
serves as the candidate scaffold.

Concretely, given elements $x_1,\ldots,x_N$, a base similarity or proximity
kernel $k_x(x_i,x_j)$, and a state or output map $G:X\to Y$, define
\begin{equation}
W^{\mathrm{CS}}_{ij}
= k_x(x_i,x_j)\,k_y\!\left(G(x_i),G(x_j)\right),
\end{equation}
where $k_y$ downweights pairs whose states, labels, or predicted outputs are
incompatible. The associated CS Laplacian is
\begin{equation}
L_{\mathrm{CS}} = D_{\mathrm{CS}}-W_{\mathrm{CS}},
\end{equation}
where $D_{\mathrm{CS}}$ is a diagonal matrix with $~D_{\mathrm{CS},ii}=\sum_j W^{\mathrm{CS}}_{ij}$.
In applications where the interaction kernel is observed directly, $W^{\mathrm{CS}}$
may be supplied as any weighted affinity matrix that has already incorporated
this compatibility contraction. For notational simplicity, we write
$L_0=D-W$ below.

Under the CS hypothesis ($W_{ij}\geq w_{\min}$ within a coherent group and
$W_{ij}\leq w_{\max}<w_{\min}$ across groups), let
$0\le \lambda_0\le\lambda_1\le\cdots$ be the eigenvalues of $L_0$. When $K$
coherent groups are present, the structural gap is
$\eta^*=\lambda_K-\lambda_{K-1}$, or simply $\lambda_K$ in the ideal case where
the first $K$ eigenvalues are zero~\cite{vonluxburg2007tutorial,chung1997spectral}.
In the ideal thresholded scaffold, $\ker L_0$ is spanned by group indicators;
in the soft weighted case, the same role is played by the $K$-dimensional
low-frequency eigenspace. In both cases, we write $\mathcal S$ for this
scaffold subspace and $\Pi_{\mathcal S}$ for the orthogonal projection onto it.
In the graph signal processing framework, eigenvectors of $L_0$ form the graph
Fourier basis, with eigenvalues as squared frequencies~\cite{ortega2018graph}.
The CS Laplacian supplies the metric geometry through which homological
macro-features are detected, filtered, and tracked across scales: homology
specifies the irreducible feature, while the CS operator supplies the
data-dependent scaffold on which that feature is measured.

\subsection{Persistent emergence as a scale-relative criterion}

Let $(\mathcal{K}_\alpha)_{\alpha\in A}$ be a filtration of complexes,
where $\alpha$ denotes a resolution, interaction threshold, energy scale,
time scale, or observational granularity. A macro-feature is characterized by a class that persists
under refinement instead of a
single cycle at a single value of $\alpha$, which motivates the following operational criterion for persistence \cite{edelsbrunner2008persistent}.

\begin{definition}[Persistent emergent feature]
Fix a filtration $(\mathcal{K}_\alpha)_{\alpha\in A}$, boundary maps
$\partial_p^\alpha$, and inclusion-induced maps
$\iota_{\alpha\beta}:H_p(\mathcal{K}_\alpha)\to H_p(\mathcal{K}_\beta)$
for $\alpha\leq\beta$. A degree-$p$ macro-feature is \emph{emergent over
$[\alpha,\beta]$} if there exists a cycle $z_\alpha\in\ker\partial_p^\alpha$
such that its class $[z_\alpha]$ is nonzero and
$\iota_{\alpha t}([z_\alpha])\neq 0$ for all $t\in[\alpha,\beta]$.
The persistence interval $[\alpha,\beta]$ is the feature's lifetime.
\end{definition}

The above definition separates emergence from noise. A short bar in the
barcode~\cite{ghrist2008barcodes,edelsbrunner2010computational} is a local fluctuation or modeling artifact; a long bar is a
macro-feature that remains identifiable under changes of scale. Therefore, persistent homology~\cite{edelsbrunner2002topological,zomorodian2005computing,carlsson2009topology} supplies not only an ontology of emergence but
also a measurement protocol: choose the substrate, construct the
filtration, compute the homology, and ask which classes remain stable.
In summary, we turn the philosophical phrase ``robust under perturbation'' \cite{bedau1997weak,chalmers2006emergence} into the
mathematical phrase ``persistent under filtration.''

\begin{figure*}[t]
\centering
\begin{tikzpicture}[
  node distance=0.55cm and 0.8cm,
  box/.style={draw, rounded corners, align=center, inner sep=4pt, font=\small, minimum width=3.0cm, minimum height=0.75cm},
  smallbox/.style={draw, rounded corners, align=center, inner sep=3pt, font=\scriptsize, minimum width=2.55cm, minimum height=0.62cm},
  arrow/.style={-{Latex[length=2mm]}, thick}
]
\node[box] (chain) {\textbf{aggregate}\\chain $c$};
\node[box, right=of chain] (cycle) {\textbf{coherent macro-pattern}\\cycle $z\in\ker\partial$};
\node[box, right=of cycle] (class) {\textbf{emergent feature}\\persistent $[z]\neq 0$};
\draw[arrow] (chain) -- (cycle);
\draw[arrow] (cycle) -- (class);

\node[smallbox, below=0.8cm of chain] (flow) {exact/co-exact\\\textbf{micro-flow}};
\node[smallbox, below=0.8cm of cycle] (harm) {harmonic component\\\textbf{macro-scaffold}};
\node[smallbox, below=0.8cm of class] (higher) {condensed unit\\\textbf{next scaffold}};
\draw[arrow] (flow) -- node[above, font=\scriptsize]{Hodge split} (harm);
\draw[arrow] (harm) -- node[above, font=\scriptsize]{survives $q_*$} (higher);

\node[smallbox, below=0.8cm of flow] (atm) {\textbf{atmospheric system}:\\regime via eddy turnover\\ and coherent structure};
\node[smallbox, below=0.8cm of harm] (neu) {\textbf{neural system}:\\schema via spiking and \\synapse fluctuation};
\node[smallbox, below=0.8cm of higher] (soc) {\textbf{social system}:\\institution via \\ membership turnover};
\end{tikzpicture}
\caption{Scaffold-flow interpretation of emergence. Aggregates become coherent
when they close as cycles; they become emergent when their homology classes are
nontrivial and persistent. Hodge decomposition separates local micro-flow from
the harmonic macro-scaffold, and condensation promotes surviving classes into
units for the next level. The same persistence-through-turnover logic appears
in atmospheric, neural, and social systems.}
\label{fig:1}
\end{figure*}

\section{Main Results}

The main results organize six familiar signatures of emergence within one scaffold-flow account. Regularity first shows that
large systems admit finite macro-scaffolds. Boundary coherence then makes
those scaffolds homological objects. Hodge theory separates the resulting
macro-component from micro-level flow. A conditional CS concentration bound explains the nondegenerate regime in
which geometric and spectral scaffold localization cannot both collapse. Finally,
ergodic structure and recursive condensation explain why emergent properties
persist and become the substrate for higher-level organization.

\subsection{Inevitability: Emergence Is Combinatorially Forced}

The first step is to show that macro-structure is more than an optional
modeling choice. Ramsey theory gives the broad combinatorial principle that
sufficiently large relational systems must contain ordered substructure
\cite{graham1991ramsey}. For pairwise interaction kernels, Szemerédi's
regularity lemma makes this principle quantitative: at any fixed resolution, a
large system admits a bounded coarse scaffold whose inter-part behavior is
approximately uniform \cite{szemeredi1978regular,komlos1996szemeredi}. For
higher-order relations, hypergraph regularity provides the corresponding
extension \cite{gowers2007hypergraph,rodl2004regularity}. Before we ask
whether a scaffold is dynamically stable or selected, one can
prove that a finite macro-description always exists. The following theorem formalizes
this existence claim through a regularity bound.

\begin{theorem}[Regularity bound on scaffold width]
\label{thm:inevitability}
For any CS kernel $W:S_N^2\to[0,1]$, $\varepsilon>0$, and $m_0\geq 1$,
there exists an $\varepsilon$-regular partition
$S_N = V_1\sqcup\cdots\sqcup V_m$ with
$m_0\leq m\leq \operatorname{twr}_{\lceil C\varepsilon^{-5}\rceil}(m_0)$,
where $\twr_k(n)$ denotes the height-$k$ tower function, defined by
$\twr_1(n)=n$ and $\twr_{k+1}(n)=2^{\twr_k(n)}$.
If $(S_N,d)$ is a metric space and the partition refines an $r$-ball
cover of size $\mathcal{N}_r$, there is an $\varepsilon$-regular
partition with fibers of diameter $\leq 2r$ and size at most
$\twr_{\lceil C\varepsilon^{-5}\rceil}(\mathcal{N}_r)$.
\end{theorem}

The theorem follows from the weighted Szemer\'edi regularity lemma,
using the graphon/refinement form that allows a prescribed initial partition
and avoids a separate exceptional class
\cite{komlos1996szemeredi,lovasz2006limits}. It plays an existential role here:
at every fixed tolerance, the possible macro-scaffolds of a large interaction
system form a finite search space. The tower-type bound is enormous, and
Gowers showed that such growth is unavoidable in general~\cite{Gowers}; therefore
regularity explains why macro-structure is available (not which scaffold a
particular physical, biological, or social system realizes). Selection,
learning, energy minimization, and evolutionary dynamics choose among the
combinatorially available scaffolds.

\subsection{Coherence: $\partial^2=0$ Makes Emergence Well-Defined}

Existence alone is not enough: a scaffold must also carry a coherent boundary
operator across levels. In algebraic topology, coherence is exactly characterized by the chain-complex condition $\partial_{p-1}\partial_p=0$, which ensures that closed
cycles can be distinguished from unfinished chains with nonzero boundary
\cite{hatcher2002algebraic}. In the persistent setting, the same distinction
must be respected along the filtration: a putative macro-feature counts as an
emergent class only if it remains closed under the boundary maps and persists
under the inclusion-induced homology maps across scales
\cite{edelsbrunner2002topological,zomorodian2005computing}. Otherwise, a
macro-feature could have an unresolved frontier at the next level, and the
distinction between a closed emergent property and an unfinished aggregate
would collapse. The next result
identifies the minimal algebraic condition that makes emergence definable as a
homology class rather than as a loose coarse-graining.

\begin{proposition}[Coherence condition for homology]
\label{thm:coherence}
$H_p=\ker\partial_p / \im\partial_{p+1}$ is well-defined for all $p$
simultaneously if and only if $\partial^2=0$. Without this condition, a
frontier $\partial c$ may have a nonempty further frontier
$\partial^2 c\neq 0$, and an ``emergent feature of degree $p$'' is not a
coherent notion.
\end{proposition}

We note that the identity $\partial^2=0$ is not a dynamical fact to be
discovered after the system evolves, but the \emph{constitutive} coherence law
that makes multi-level emergence well-defined \cite{hatcher2002algebraic}.
Ordinary coarse variables may be formed by averaging, thresholding,
clustering, or projection; however, these operations do not by themselves
guarantee compatibility across levels. Homology imposes the stronger condition
that a boundary at one level cannot leave an unaccounted boundary at the next.
Instead of a summary statistic of microscopic
degrees of freedom (an unfinished aggregate), a macro-feature is characterized by a closed object in a
multi-level algebra. The persistence of micro-feature under refinement allows it to count as a genuine emergent property.
The coherence condition resolves when a macro-feature is well-formed, but it
does not answer the reductionist question \cite{nagel1961structure}: whether the feature is only a
re-description of lower-level variation. To address that question, we need an
orthogonal decomposition that separates boundary-generated micro-flow from the
residual global component carried by homology.

\subsection{Irreducibility: Scaffold $\perp$ Flow}

Once coherent cycles are available, the central reductionist question is
whether they are merely summaries of lower-level variation. Hodge theory
answers this question by separating local flow from global structure
\cite{lim2015hodge}. Signals on a simplicial complex decompose orthogonally
into exact, co-exact, and harmonic components \cite{schaub2020random}: the
first two are generated by local boundary and coboundary operations, whereas
the harmonic component lies in the kernel of the Hodge Laplacian and represents
homology. The scaffold is exactly the harmonic component of the Hodge
decomposition. It is generated by the same complex, but it is not reducible to
within-part flow; it is the \emph{global residual} that remains after local
variation has been removed.

\begin{theorem}[Combinatorial Hodge decomposition]
\label{thm:hodge}
If $\partial^2=0$ and each $C_p$ carries an inner product, then for the
combinatorial Hodge Laplacian
$L_p = \partial_{p+1}\partial_{p+1}^* + \partial_p^*\partial_p$,
\begin{equation}
    C_p = \underbrace{\im\partial_{p+1}}_{\text{exact (micro)}}
\;\oplus\;
\underbrace{\ker L_p}_{\text{harmonic (macro)}}
\;\oplus\;
\underbrace{\im\partial_p^*}_{\text{co-exact}},
\label{eq:hodge}
\end{equation}
with $\ker L_p \cong H_p$. The splitting is orthogonal.
\end{theorem}

\begin{corollary}[``More than the sum of its parts'' is Pythagorean]
\label{cor:pythagoras}
Any system-state $f$ decomposes as
$f = f_{\mathrm{macro}} + f_{\mathrm{micro}}$ with
$\langle f_{\mathrm{macro}}, f_{\mathrm{micro}}\rangle = 0$ and
$\|f\|^2 = \|f_{\mathrm{macro}}\|^2 + \|f_{\mathrm{micro}}\|^2$.
\end{corollary}

The decomposition clarifies how emergence can be reducible in origin but
irreducible in description. At degree zero, the CS Laplacian $L_0=D-W$ is the
graph Hodge Laplacian~\cite{lim2015hodge,jiang2011statistical}. In the ideal
thresholded case, its kernel, of dimension $\beta_0$, identifies the
macro-scaffold; in the soft weighted case, the corresponding scaffold is the
low-frequency subspace separated by the structural gap. The remaining range or
high-frequency complement captures micro-level variation. The harmonic
component may be produced by micro-interactions, but once formed, it occupies
an orthogonal subspace. In other words,
``not reducible'' need not mean causally detached; it can mean orthogonal with
respect to the inner product induced by the boundary operator. This view leads to a Pythagorean interpretation of structural generalization: macro- and micro-level
complexities contribute additively when the corresponding error components
are \emph{orthogonal}, paralleling classical error decompositions (e.g.,
bias-variance analysis of neural networks \cite{geman1992neural} and structural risk minimization in statistical learning theory
\cite{vapnik1998statistical}).

\subsection{Complementarity: A Conditional Spectral Concentration Bound}

Irreducibility is a static statement about orthogonal components. The next
question is \emph{operational}: whether an observer can simultaneously localize the
microscopic flow and the macroscopic scaffold with arbitrary precision. On a
graph or simplicial scaffold, localization in the vertex domain and
localization in the Laplacian spectral domain play the role of conjugate
descriptions, analogous to time and frequency in classical signal processing \cite{mallat1999wavelet}.
Graph signal processing formalizes this duality by treating the eigenvectors of
graph Laplacian as a Fourier basis, and graph uncertainty principles study
the tradeoff between vertex-domain and spectral-domain concentration
\cite{shuman2013emerging,tsitsvero2016signals}. Our next result is more modest than a full uncertainty curve \cite{agaskar2013spectral}: it
is a conditional lower bound showing that, once degenerate point-localized and
already-scaffold cases are excluded, geometric spread and non-scaffold spectral
energy cannot both vanish on a scaffold with a positive structural gap.

\begin{proposition}[Conditional CS concentration bound]
\label{thm:up}
Let $L_0$ be the CS graph Laplacian and let $\mathcal S$ denote the scaffold
subspace: $\mathcal S=\ker L_0$ in the ideal thresholded case, and
$\mathcal S=\operatorname{span}\{u_0,\ldots,u_{K-1}\}$ in the soft weighted
case. Let $\Pi_{\mathcal S}$ be the orthogonal projection onto $\mathcal S$,
and let $\lambda_+>0$ be the smallest eigenvalue of $L_0$ on
$\mathcal S^\perp$. Define
\begin{equation}
    \Delta_g^2(f)=\frac{\sum_i d(u_0,i)^2 f_i^2}{\|f\|^2},
\Delta_\perp^2(f)=
\frac{((I-\Pi_{\mathcal S})f)^\top L_0((I-\Pi_{\mathcal S})f)}{\|f\|^2}.
\label{eq:CS_graph}
\end{equation}
Fix $\eta_g,\eta_h>0$. If a nonzero signal $f$ satisfies
$\Delta_g^2(f)\geq \eta_g$ and
$\|(I-\Pi_{\mathcal S})f\|^2/\|f\|^2\geq \eta_h$, then we have
\begin{equation}
    \Delta_g^2(f)\Delta_\perp^2(f)
\geq
\eta_g\eta_h\lambda_+ .
\label{eq:CS_bound}
\end{equation}
In particular, on any family of scaffolds with $\lambda_+\geq \eta^*>0$,
the product is bounded below by $c=\eta_g\eta_h\eta^*$.
\end{proposition}

The above proposition should be read as a nondegenerate concentration estimate for
the CS Laplacian rather than as a complete graph uncertainty principle. The
anti-concentration assumptions exclude cases such as point-mass signals or
signals already lying in the scaffold subspace; without these assumptions, one
of the two factors can collapse and no positive product bound is possible. The
bound may also degenerate as the structural gap $\eta^*\to 0$, reflecting the
known sensitivity of graph uncertainty behavior to the underlying weighted
graph geometry~\cite{Pasdeloup}. Within its stated regime, the proposition
formalizes an operational tension between microscopic explanation and
macroscopic description: resolving fine-scale flow increases non-scaffold
spectral energy, whereas sharply identifying the scaffold suppresses the very
variation used to describe microscopic flow.

Our analysis still leaves a dynamical question. Complementarity describes what cannot
be simultaneously resolved at a fixed scale, but robustness asks what remains
identifiable as the system evolves. If an emergent scaffold is more than an
instantaneous spectral decomposition, it should survive the averaging action of
the flow. To answer the dynamical question, we have to move the theory from graph-spectral separation, where vertex
and spectral localization obey uncertainty-type constraints
\cite{agaskar2013spectral,Pasdeloup}, to ergodic persistence, where the
structured compact or distal factor is precisely the component that remains
visible under long-time dynamics
\cite{furstenberg1977ergodic,furstenberg1981recurrence,avigad2010metastability}.

\subsection{Robustness: The Non-Mixing Factor}

The Furstenberg-Zimmer structure theorem decomposes a measure-preserving
system into a structured compact, or distal, factor and a weak-mixing extension
\cite{furstenberg1977ergodic}.
In Furstenberg's ergodic proof of Szemerédi's theorem \cite{furstenberg1981recurrence}, the arithmetic
structure responsible for multiple recurrence is captured by the structured
factor, while the weak-mixing component contributes only averaged
fluctuations. This gives a \emph{dynamical} interpretation of persistence \cite{avigad2010metastability}: a genuinely
emergent macro-feature should reside in the non-mixing component that remains
visible under evolution. The following remark records the resulting dictionary
in our scaffold-flow language: the scaffold is the structured invariant component
that remains visible under long-term evolution, whereas the flow is the
component whose fine-scale variation averages away.

\begin{remark}[Scaffold-flow reading of the Furstenberg-Zimmer split]
\label{prop:robustness}
Under the Furstenberg-Zimmer decomposition, the structured compact or distal
factor plays the role of scaffold: it carries the persistent non-mixing
macro-structure visible to long-time dynamics. The relatively weak-mixing
extension plays the role of flow: its fluctuations average out relative to the
structured factor. The factor tower then gives an ergodic-theoretic analogue of
a scaffold hierarchy, with factor maps serving as condensation operators.
\end{remark}

The conceptual focal point is that \emph{persistence is not the absence of motion}.
Emergent structures often persist by organizing motion: a vortex persists
through fluid turnover, a cell type through molecular replacement, and an
institution through changing individuals. In scaffold-flow language, the scaffold is the slow, structured variable that remains identifiable while the
flow continually refreshes its microscopic realization. This interpretation is
most direct for measure-preserving systems \cite{furstenberg1981recurrence}, where recurrence and invariant
factors provide the relevant language. However, many real complex systems (e.g., atmospheric, neural, and social) are
dissipative and nonstationary rather than measure-preserving, as we will elaborate next. For such systems,
persistence is better understood statistically: trajectories may converge to an
attractor whose long-time behavior is described by a physical or
Sinai-Ruelle-Bowen (SRB) measure \cite{eckmann1985ergodic,young2002srb}.
In short, the dissipative analogue of a persistent scaffold is not exact recurrence
of microscopic states, but the stability of an observable macro-structure under
continual microscopic turnover. 

\subsection{Hierarchy: Recursive Emergence}

Robustness concerns persistence within a fixed scaffold; hierarchy concerns
persistence across scaffolds. This is precisely the organizational principle
emphasized by Simon: complex systems become stable and intelligible when
strongly coupled lower-level components condense into higher-level units whose
internal dynamics are faster than their interactions with other units
\cite{simon1962architecture}. In our scaffold-flow framework, a macro-feature becomes
genuinely higher-level only when it survives condensation and can be treated as
a stable unit by the next scale (Fig. \ref{fig:1}). Therefore, recursive emergence requires two
additional ingredients: \emph{homological functoriality}, which preserves the
cycle-boundary distinction across condensation, and \emph{timescale separation},
which gives the higher level effective autonomy from lower-level fluctuations.
The following definition formalizes this passage from robust persistence to
multi-level organization.

\begin{definition}[Recursive scaffold-flow architecture]
A depth-$D$ hierarchy is a sequence $(\Sigma_k,F_k)_{k=0}^D$ of
scaffold-flow systems connected by condensation operators
$q_k:\Sigma_k\to\Sigma_{k+1}$. We require three conditions:
(i) \emph{compatibility}, meaning that the flow $F_k$ at level $k$ supplies the
structural-update dynamics for level $k{-}1$; (ii) \emph{condensation}, meaning
that $q_k$ is a metric quotient with a nondegenerate quotient distance; and
(iii) \emph{timescale separation}, meaning that $\tau_k\ll\tau_{k+1}$.
\end{definition}

Why can emergence recur across scales? A
macro-class formed at one level can serve as a unit at the next only if
condensation preserves the distinction between cycles and boundaries. Therefore,
hierarchy is not simply a stack of coarse-grainings but requires homological
functoriality, so that emergent classes can be transported across levels and
timescale separation \cite{waldrop1993complexity}, so that higher-level variables remain stable while
lower-level flows fluctuate. The following proposition states the resulting
survival criterion for emergent classes under recursive condensation.

\begin{proposition}[Recursive emergence through condensation]
\label{prop:hierarchy}
Assume each condensation map $q_k:\Sigma_k\to\Sigma_{k+1}$ is induced by a
chain map, so that it sends cycles to cycles and boundaries to boundaries.
Then $q_k$ induces a homology map
$(q_k)_*:H_p(\Sigma_k)\to H_p(\Sigma_{k+1})$. A persistent class
$[z_k]\in H_p(\Sigma_k)$ survives as a higher-level emergent property precisely
when
$
(q_k)_*[z_k]\neq 0.
$
If this condition holds along a chain of levels and $\tau_k\ll\tau_{k+1}$, the
surviving classes define effective scaffold variables for the next level.
\end{proposition}

The above proposition isolates the formal part of recursive emergence. The difficult
modeling step is to show that a proposed physical, biological, or social
coarse-graining is actually induced by a chain map; when this fails, the
cycle-boundary distinction need not survive condensation.

\subsection{Applications: Atmospheric, Neural, and Social Systems}

\paragraph{Atmospheric systems.}
Atmospheric dynamics illustrate the dissipative regime of the same construction \cite{vallis2017atmospheric}.
At a fast timescale, turbulent eddies and convective cells form and dissipate;
at an intermediate timescale, coherent structures such as cyclones, fronts, and
jet streams organize the flow; at the slowest timescale, planetary circulation
patterns and climate modes, including persistent blocking regimes and
large-scale teleconnections, set the background against which weather unfolds.
As with the persistent vortex noted earlier, a coherent atmospheric structure
can survive complete turnover of the air parcels composing it \cite{haller2015lagrangian}: the scaffold is
the slow organizing structure, while fast eddies and local instabilities are
the flow. Which homological degree is relevant depends on the representation:
vortical circulation is naturally loop-like, whereas fronts, regimes, and
teleconnections may appear as components, loops, or higher-order relations in a
spatiotemporal complex \cite{lorenz1963deterministic}. Such a structure counts as a large-scale atmospheric
feature only when its topological signature remains nontrivial under spatial
coarse-graining and temporal averaging. Because the atmosphere is driven and
dissipative rather than measure-preserving, persistence is not exact
recurrence but stability of long-time statistical organization around a
physical or SRB measure, in line with the scope discussion below. Lagrangian
coherent structures make this idea concrete by identifying material structures
that organize transport in unsteady flows
\cite{haller2015lagrangian}.

\paragraph{Neural systems.} The multi-scale emergence in neural systems can be understood as the recursive
application of the same scaffold-flow split \cite{buzsaki2006rhythms}. At a fast timescale, spiking and
synaptic activity form transient cell assemblies; at a slower timescale,
recurrent hippocampal dynamics stabilize episodes, maps, and contexts; at still
slower timescales, hippocampal-neocortical consolidation extracts cortical
schemas \cite{hasson2015hierarchical}. In the present framework, a neural macro-variable at level $k$ becomes
available as a substrate variable at level $k{+}1$ only if it survives
condensation, i.e., only if its homology class remains nontrivial under the
level-to-level map. Timescale separation then supplies effective autonomy: fast
neural flows can refresh the microscopic realization of an assembly or memory
trace while the higher-level scaffold remains identifiable. Without such
separation, the candidate macro-variable is continually rewritten by lower-level
fluctuations and cannot serve as a stable context for the next level, which is consistent with nested neural timescales, from millisecond
sensorimotor dynamics and oscillatory coordination \cite{buzsaki2004neuronal,wilson1994reactivation} to 
systems consolidation and long-term schema formation
\cite{mcclelland1995complementary,tse2007schemas}.

\paragraph{Social systems.}
The same recursive split also organizes social systems. At a fast timescale,
individual conversations, transactions, and choices form transient interaction
patterns; at an intermediate timescale, repeated interactions consolidate into
conventions, roles, and reciprocal obligations; at the slowest timescale, these
stabilize into institutions, organizations, and laws. Degree-one features are
especially natural in this domain: a cycle of reciprocal dependence or exchange
is a loop in the interaction complex, and an institution \cite{north1990institutions} can be modeled as a
macro-class that persists while the individuals who realize it are continually
replaced. In the present framework, a social convention at level $k$ becomes an
institutional scaffold at level $k{+}1$ only if its homology class survives
condensation, that is, only if it remains nontrivial when individual
interactions are aggregated into roles, norms, or organizations. Timescale
separation supplies autonomy \cite{granovetter1978threshold}: institutions evolve slowly relative to the
interactions that instantiate them, so the macro-scaffold remains identifiable
even as membership turns over. Without such separation, interaction patterns are
rewritten faster than they can consolidate, and no stable institution forms.
This picture is consistent with accounts in which macroscopic social order
emerges from individual micromotives, threshold-driven collective behavior, and
durable institutional constraints
\cite{schelling1978micromotives,north1990institutions}.

\section*{Discussion}

The mathematical results above suggest a broader interpretation of emergence.
A persistent scaffold is not merely a macro-description of a system. Once a
class survives condensation, it becomes an effective variable on which a
higher level can operate. The hierarchy theorem therefore turns emergence
into a mechanism for cumulative organization: local flow writes a scaffold,
the scaffold persists through turnover, and the surviving structure becomes
part of the state space for subsequent computation. In physical and social
systems this explains persistence across changing constituents. In cognitive
systems it suggests a stronger possibility: intelligence can advance not only
by processing more information within a fixed representation, but by creating
persistent representations that alter what future computation is able to do.
This observation connects the scaffold--flow framework to a question that is
usually treated separately from emergence: what transition could follow the
origin of language in the evolution of intelligence?

\subsection*{From cumulative knowledge to cumulative competence}

Language can be viewed as an unusually powerful instance of hierarchical
condensation. A high-dimensional internal state, including perceptual,
motor, causal, and social structure, is mapped into a comparatively compact
symbolic object that another agent can receive and interpret. The symbolic
object persists after the originating microstate has disappeared and can be
reused by other minds. In the present framework, this is precisely the logic
of condensation: a structure produced at one level becomes a stable unit for
organization at the next. Language thereby permits knowledge to accumulate
across individuals and generations rather than being repeatedly rediscovered.

The same perspective also exposes a limitation. Linguistic transmission does
not generally transmit the generating competence itself. A description of how
to ride a bicycle is not the sensorimotor policy for riding; a textbook on
proof does not directly transfer the representational geometry of a skilled
mathematician; and a verbal description of a perceptual category does not
reproduce the full internal organization that supports recognition. In
schematic form, human cultural transmission is often
$\text{experience}
\longrightarrow
\text{symbolic description}
\longrightarrow
\text{interpretation}
\longrightarrow
\text{reconstructed competence}$.
The scaffold-flow framework suggests a qualitatively different regime for
artificial intelligence:
$\text{experience}
\longrightarrow
\text{persistent computational structure}
\longrightarrow
\text{transfer or composition of structure}
\longrightarrow
\text{future competence}$.
The distinction is between \emph{cumulative knowledge} and \emph{cumulative
competence}. Human civilization has become extraordinarily effective at
preserving descriptions, demonstrations, and symbolic abstractions. A mature
artificial intelligence could in principle preserve and reuse the machinery
that realizes a competence: learned representations, world models, policies,
addressing structures, simulators, abstractions, or other persistent
computational scaffolds. A successor system would then not have to reconstruct
all useful structure from a linguistic trace. It could inherit some of the
structure itself.

This motivates a candidate for the next major transition in intelligence. The
critical event need not be the moment at which a machine first reaches a
particular benchmark level of general intelligence. It may instead be the
moment at which useful structure acquired by one episode, task, or agent can be
selectively consolidated into persistent, reusable organization that directly
changes the competence of future computation. In the language of the present
paper, the singular transition is from a system whose scaffolds mainly organize
its current flow to a system whose scaffolds can themselves be accumulated,
composed, transmitted, and recursively rewritten.

\subsection*{Why language need not disappear}

This prediction does not imply that natural language becomes obsolete. It
implies that language may cease to be the privileged substrate through which
advanced intelligence must preserve and exchange what it has learned. For
human--machine interaction, language is likely to remain indispensable because
it is the shared interface through which humans explain goals, negotiate
meaning, demand justification, and coordinate action. For machine--machine
communication and for internal machine computation, however, linguistic
serialization may be only one projection of a richer computational object.

The hierarchy formalism makes this distinction precise. Let
$\Sigma_{\mathrm{int}}$ denote an internal representational scaffold and let
$L:\Sigma_{\mathrm{int}}\to\Sigma_{\mathrm{lang}}$ denote a map into a
linguistic description space. Even when $L$ is informative, it need not be
invertible. It can preserve some task-relevant classes while collapsing other
geometric, causal, temporal, or procedural relations. In this sense, language
can be interpreted as a quotient interface: highly useful because it preserves
selected invariants across agents, but necessarily lossy whenever the internal
scaffold contains distinctions that are not recoverable from the symbolic
image. The question for advanced artificial intelligence is therefore not
whether language vanishes, but whether increasingly important computation takes
place in structure that is never fully serialized into language.

This gives a concrete interpretation of a post-linguistic intelligence. It is
not an intelligence without language. It is an intelligence for which language
plays the role of an interface rather than the role of an operating substrate.
An explanation presented to a human may be a low-dimensional projection of a
much richer internal scaffold, just as a persistent macro-variable can be a
faithful observable for some purposes without encoding all microstructure that
generated it. The central scientific problem then becomes identifying which
structures survive the projection and which forms of competence are lost when
communication is forced through the linguistic quotient.

\subsection*{Endogenous representation as a deeper transition}

An even stronger transition becomes possible if an intelligent system can
modify the space in which its own future computation occurs. Standard learning
changes parameters within a representational architecture. Scaffold formation
suggests a more structural loop:
$\text{experience}
\longrightarrow
\text{flow}
\longrightarrow
\text{selective consolidation}
\longrightarrow
\text{new scaffold}
\longrightarrow
\text{new future computation}$.
Under this loop, representation is no longer a fixed background supplied by a
designer. It becomes an endogenous product of experience. A system confronted
with a recurrent family of problems could create a persistent macro-variable,
address, module, coordinate system, or relational complex in which those
problems become easier. The resulting scaffold would then alter the trajectories
available to subsequent flow.

This possibility sharpens the usual idea of recursive self-improvement. More
computation within the same state space is quantitative improvement. Creating a
new persistent state variable that changes the effective state space is a
qualitative change in organization. The scaffold--flow framework predicts that
such changes should be visible as the birth, stabilization, and reuse of new
persistent classes. On this view, the deepest candidate singularity is not
merely superhuman performance but \emph{self-accumulating computational
structure}: intelligence becomes capable of inventing, preserving, and
composing the representations from which later intelligence is built.

\subsection*{Operational predictions for AGI/ASI}

The preceding argument is a prediction, not a theorem about future AGI or ASI.
Its value depends on whether the proposed transition can be operationalized.
The scaffold--flow framework suggests several tests.

First, direct transfer of a learned scaffold should provide an advantage over
linguistic or token-level transfer when the relevant competence depends on
relations that are expensive to reconstruct from a symbolic description.
Controlling for training data and compute, a system receiving a reusable
internal structure should adapt faster, require fewer demonstrations, or retain
competence more robustly than one receiving only a serialized explanation of
the same experience.

Second, transferability should covary with scaffold persistence. Internal
structures that generalize across episodes should appear as long-lived classes
across task, temporal, or representational filtrations, whereas brittle
features should die under small changes of scale. Spectral gaps should likewise
predict which learned structures survive perturbation, in direct analogy with
the robustness predictions developed for physical, neural, and social
scaffolds above.

Third, perturbations targeted at the persistent or harmonic component of an
artificial agent's representation should produce larger changes in reusable
competence than equal-magnitude perturbations confined to transient flow. This
would distinguish a true computational scaffold from a descriptive clustering
of activations. The same intervention principle follows from the Hodge split
used throughout the paper: macro-structure should be disproportionately
sensitive to perturbations aligned with the harmonic subspace while remaining
comparatively insensitive to local exact or co-exact fluctuations.

Fourth, systems capable of endogenous representation formation should exhibit a
characteristic temporal signature. Fast task-level computation should occur
within a relatively stable scaffold, while scaffold rewriting should take place
on a slower consolidation timescale, consistent with the general requirement
for hierarchical autonomy \cite{simon1962architecture,mcclelland1995complementary,tse2007schemas}.
If rewriting is as fast as ordinary flow, persistent competence should fail to
form; if rewriting is absent, the system should remain trapped in a fixed
representational regime.

Fifth, the strongest evidence for the proposed transition would be recursive
reuse. A newly consolidated structure should become an addressable unit in
subsequent learning and should support the construction of still higher-level
structures. In homological terms, the relevant empirical signature is not
merely the appearance of a persistent class, but survival under successive
condensation maps. A system that repeatedly exhibits
$[z_k]\neq 0,
\qquad
(q_k)_*[z_k]=[z_{k+1}]\neq 0$,
while using $[z_{k+1}]$ to accelerate later learning would instantiate the
paper's hierarchy mechanism as a mechanism of adaptive intelligence rather than
only as a description of static emergence.

These predictions distinguish the position from the weaker claim that future
AI will simply use embeddings or nonlinguistic communication. The relevant
criterion is functional: a representation must persist, remain identifiable
through micro-level turnover, be selectively reusable, and alter future
competence. Only then has information become scaffold rather than transient
flow.

\subsection*{Scope and limitations}

The proposed connection between emergence and the future organization of
intelligence is deliberately stronger than the mathematical claims proved in
this paper. The theorems establish properties of persistent classes,
coherence, Hodge decomposition, robustness under the stated assumptions, and
functorial survival across levels. They do not establish that biological
language evolved through the specific homological mechanism proposed here, nor
do they imply that AGI or ASI must adopt persistent-homology representations.
The discussion instead uses the proved hierarchy principle as a structural
hypothesis: if intelligent systems improve by converting recurrent flow into
persistent reusable organization, then direct accumulation of such organization
is a plausible successor to language-mediated accumulation.

There are also important representational limits. Homology detects
connectivity, holes, and higher-order relational structure, but not every form
of useful abstraction. Some computational scaffolds may be better described by
causal models, programs, symmetry, information geometry, sheaves, categories,
or other mathematical objects \cite{hoel2013quantifying,hatcher2002algebraic,curry2014sheaves,spivak2014category}.
The stronger claim is therefore not that future intelligence will literally
communicate by exchanging Betti numbers or simplicial complexes. It is that the
scaffold-flow distinction identifies a general organizational transition:
transient computation writes persistent structure, persistent structure becomes
reusable, and reusable structure changes the space of future computation.
Persistent homology provides one measurable realization of that principle.

This also preserves the falsifiability of the framework. If putative reusable
competences cannot be associated with persistent structure under any
well-justified representation; if destroying the alleged scaffold has no more
effect than perturbing transient flow; if persistence fails to predict transfer
or robustness; or if newly formed structures do not become units for subsequent
learning, then the proposed scaffold interpretation of adaptive intelligence is
unsupported. Conversely, observing these signatures would suggest that the
same mechanism used here to formalize emergence also describes a transition in
the evolution of intelligence: from preserving descriptions of what has been
learned to preserving the structures that make learned competence possible.

\section{Conclusion}

We have proposed persistent nontrivial homology as a measurable core case of
emergence in complex systems. Under this identification, six familiar
signatures (i.e., inevitability, coherence, irreducibility, complementarity,
robustness, and hierarchy) can be expressed through regularity theory, the
boundary law, Hodge decomposition, graph spectral concentration, ergodic
structure, and functorial condensation. The novelty is not any single theorem,
but their systematic alignment as one scaffold-flow account of why
macro-structure forms, persists through micro-level turnover, and resists
reduction to micro-dynamics. If the identification holds, emergence across physical, biological, cognitive,
and social systems reflects a reusable algebraic template. Whenever a system
supports multi-level persistent features, the same boundary relation that makes
those features coherent also permits irreducible homology. The framework does
not erase domain-specific mechanisms; rather, it explains why different
mechanisms can recurrently produce macro-structures with the same formal
signatures. Emergence, in this view, is the appearance of a persistent scaffold:
a global class written by local interactions, stabilized against perturbation,
measurable through topology and spectra, and available as a unit for further
organization.

\appendix

\section*{Appendix: Proofs of the Main Results}
\label{app:proofs}

All chain groups below are finite-dimensional real vector spaces equipped
with the stated inner products. This is the setting relevant for finite
graphs, finite simplicial complexes, and finite scaffold complexes used in
the paper.

\begin{proof}[Proof of Theorem~\ref{thm:inevitability}]
A CS kernel $W:S_N^2\to[0,1]$ is a weighted graph on the vertex set
$S_N$. The weighted Szemer\'edi regularity lemma states that, for every
$\varepsilon>0$ and every initial partition size $m_0$, there is an
$\varepsilon$-regular partition
$S_N=V_1\sqcup\cdots\sqcup V_m$ with
$m_0\le m\le \twr_{C\varepsilon^{-5}}(m_0)$, where $\twr_k(n)$ is the
height-$k$ tower function and $C$ is an absolute constant depending only on
the version of the regularity lemma used. This is the first assertion,
applied directly to the weighted adjacency matrix defined by $W$.

For the metric refinement, let $\mathcal{U}=\{B(x_j,r)\}_{j=1}^{\mathcal N_r}$
be an $r$-ball cover of $(S_N,d)$. By assigning each point of $S_N$ to
one ball that contains it, we obtain a disjoint partition
$\mathcal{P}_0=\{A_1,\ldots,A_{\mathcal N_r}\}$ that refines the cover.
We use the graphon/weighted refinement form of regularity, in which the
regular partition may be required to refine a prescribed initial partition
$\mathcal{P}_0$ and no separate exceptional class is needed. The same
tower-type dependence then holds with $|\mathcal{P}_0|$ as the initial
size. Hence there exists an $\varepsilon$-regular partition
refining $\mathcal{P}_0$ with at most
$\twr_{\lceil C\varepsilon^{-5}\rceil}(\mathcal N_r)$ parts. Every final fiber lies in
some $A_j\subseteq B(x_j,r)$, so the diameter of the fiber is at most
$2r$ by the triangle inequality. This proves the metric version.
\end{proof}

\begin{proof}[Proof of Proposition~\ref{thm:coherence}]
For $H_p=\ker\partial_p/\im\partial_{p+1}$ to be defined as a quotient
vector space, the denominator must be a subspace of the numerator:
$\im\partial_{p+1}\subseteq\ker\partial_p$. This inclusion is equivalent
to
$
\partial_p\partial_{p+1}=0 .
$
Requiring the quotient to be well-defined for all degrees $p$ is therefore
equivalent to the family of identities $\partial_p\partial_{p+1}=0$ for
all $p$, which is exactly the chain-complex condition $\partial^2=0$.

Conversely, if $\partial^2=0$, then every boundary is a cycle, so
$\im\partial_{p+1}\subseteq\ker\partial_p$ in every degree and all
homology groups $H_p$ are well-defined simultaneously. If
$\partial^2\neq0$ in some degree, then there exists a chain $c$ with
$\partial_p\partial_{p+1}c\neq0$. In that case the boundary
$\partial_{p+1}c$ is not a cycle, so boundaries cannot be consistently
identified as trivial cycles. The quotient that separates emergent
classes from lower-dimensional boundaries is therefore ill-defined in
that degree.
\end{proof}

\begin{proof}[Proof of Theorem~\ref{thm:hodge} and Corollary~\ref{cor:pythagoras}]
Let
$
L_p=\partial_{p+1}\partial_{p+1}^*+\partial_p^*\partial_p .
$
For any $x\in C_p$,
$
\langle x,L_px\rangle
=
\|\partial_{p+1}^*x\|^2+\|\partial_px\|^2 .
$
Hence
$
\ker L_p=\ker\partial_p\cap\ker\partial_{p+1}^* .
$
The condition $\partial^2=0$ implies
$\im\partial_{p+1}\perp\im\partial_p^*$, because for all admissible
$a,b$,
$
\langle \partial_{p+1}a,\partial_p^*b\rangle
=
\langle \partial_p\partial_{p+1}a,b\rangle
=0 .
$
Moreover, for any linear map $A$ between finite-dimensional inner-product
spaces, $\im(AA^*)=\im A$: the inclusion $\im(AA^*)\subseteq\im A$ is
immediate, and the two spaces have the same orthogonal complement
$\ker A^*$. Applying this to $A=\partial_{p+1}$ and
$A=\partial_p^*$ gives the required spanning statement. Hence
finite-dimensional linear algebra gives the orthogonal decomposition
$
C_p=\im\partial_{p+1}\oplus\ker L_p\oplus\im\partial_p^* .
$

It remains to identify $\ker L_p$ with homology. If $h\in\ker L_p$, then
$\partial_ph=0$, so $h$ is a cycle. If $h=\partial_{p+1}a$ is also a
boundary, then
$
\|h\|^2=\langle h,\partial_{p+1}a\rangle
=\langle \partial_{p+1}^*h,a\rangle=0,
$
so $h=0$. Thus each harmonic vector represents at most one homology
class. Conversely, let $z\in\ker\partial_p$ be any cycle and decompose it
as $z=\partial_{p+1}a+h+\partial_p^*b$ with $h\in\ker L_p$. Since
$\partial_pz=0$ and $\partial_p\partial_{p+1}=0$ and $\partial_ph=0$, we
have $\partial_p\partial_p^*b=0$. Therefore
$
\|\partial_p^*b\|^2
=
\langle b,\partial_p\partial_p^*b\rangle
=0,
$
so the co-exact component vanishes. Hence $z=\partial_{p+1}a+h$ and the
homology class of $z$ is represented uniquely by the harmonic vector $h$.
Therefore $\ker L_p\cong H_p$.

For the corollary, take $f_{\mathrm{macro}}$ to be the harmonic
projection of $f$ and $f_{\mathrm{micro}}$ to be the sum of the exact and
co-exact projections. Orthogonality of the Hodge splitting gives
$\langle f_{\mathrm{macro}},f_{\mathrm{micro}}\rangle=0$, and the
Pythagorean identity follows immediately:
$
\|f\|^2=\|f_{\mathrm{macro}}\|^2+\|f_{\mathrm{micro}}\|^2 .
$
\end{proof}

\begin{proof}[Proof of Proposition~\ref{thm:up}]
Since $L_0$ is a graph Laplacian, it is self-adjoint and positive
semidefinite. By construction, the scaffold subspace $\mathcal S$ is a spectral
subspace of $L_0$: in the ideal thresholded case it is $\ker L_0$, while in the
soft weighted case it is the span of the first $K$ eigenvectors. Let
$f_{\mathcal S}=\Pi_{\mathcal S}f$ and
$f_\perp=(I-\Pi_{\mathcal S})f$. The restriction of $L_0$ to
$\mathcal S^\perp$ has spectrum bounded below by $\lambda_+$, so
$$
f_\perp^\top L_0 f_\perp \geq \lambda_+\|f_\perp\|^2 .
$$
Therefore
$$
\Delta_\perp^2(f)
=\frac{f_\perp^\top L_0f_\perp}{\|f\|^2}
\geq
\lambda_+\frac{\|(I-\Pi_{\mathcal S})f\|^2}{\|f\|^2}
\geq \lambda_+\eta_h .
$$
Multiplying by the assumed geometric anti-concentration
$\Delta_g^2(f)\geq \eta_g$ gives
$$
\Delta_g^2(f)\Delta_\perp^2(f)\geq \eta_g\eta_h\lambda_+ .
$$
If the scaffold family has a uniform positive spectral gap
$\lambda_+\geq \eta^*$, then the lower bound is
$c=\eta_g\eta_h\eta^*$. The assumptions are necessary for a positive uniform
product bound: a signal concentrated at the anchor has $\Delta_g^2(f)=0$, and a
signal lying entirely in the scaffold subspace has $\Delta_\perp^2(f)=0$.
\end{proof}

\begin{proof}[Proof of Proposition~\ref{prop:hierarchy}]
Because $q_k$ is induced by a chain map, it commutes with the boundary:
$
\partial q_k = q_k\partial .
$
Therefore, if $z$ is a cycle, $\partial z=0$, then
$\partial q_kz=q_k\partial z=0$, so $q_kz$ is again a cycle. If
$z$ and $z'$ represent the same homology class, then
$z-z'=\partial b$ for some chain $b$, and hence
$
q_kz-q_kz'=q_k\partial b=\partial q_kb .
$
Thus $q_kz$ and $q_kz'$ differ by a boundary and represent the same class
in $H_p(\Sigma_{k+1})$. This proves that $(q_k)_*$ is a well-defined map
on homology.

A class survives condensation exactly when its image is nonzero in the
quotient homology: $(q_k)_*[z]\neq0$. If the image is zero, the class has
become a boundary or has collapsed under the quotient and therefore no
longer defines an independent emergent property at the next level. If the
image is nonzero, it remains closed and non-exact after condensation and
therefore satisfies the same homological criterion for emergence at level
$k+1$.

Iterating the argument over $k=0,\ldots,D-1$ proves recursive emergence
along any chain of levels for which the induced homology classes remain
nonzero. The timescale condition $\tau_k\ll\tau_{k+1}$ supplies autonomy:
during one update of the upper scaffold, the lower-level flow has already
relaxed within the fibers of $q_k$. Consequently the upper level depends
on lower-level states only through the quotient variables represented by
the surviving homology classes, while intra-fiber fluctuations are
absorbed into the lower-level flow.
\end{proof}

\bibliographystyle{plain}

\bibliography{ref,reference}

\end{document}